\documentclass[11pt]{article}

\usepackage[T1]{fontenc}
\usepackage[utf8]{inputenc}
\IfFileExists{lmodern.sty}{\usepackage{lmodern}}{}
\usepackage[activate={true,nocompatibility},expansion=false]{microtype}

\usepackage[letterpaper,top=1in,bottom=1in,left=1.4in,right=1.4in]{geometry}
\usepackage{setspace}

\usepackage{amsmath,amssymb,amsfonts}
\usepackage{graphicx}
\usepackage{booktabs}
\usepackage{array}
\usepackage{caption}
\usepackage{xcolor}
\definecolor{linkblue}{RGB}{0,60,140}
\usepackage[colorlinks=true,linkcolor=linkblue,citecolor=linkblue,urlcolor=linkblue]{hyperref}
\usepackage{url}

\usepackage{titlesec}
\titleformat{\section}{\normalfont\large\bfseries}{\thesection}{0.75em}{}
\titleformat{\subsection}{\normalfont\normalsize\bfseries}{\thesubsection}{0.6em}{}
\titlespacing*{\section}{0pt}{1.4em}{0.5em}
\titlespacing*{\subsection}{0pt}{1.0em}{0.35em}

\newcommand{\LN}{\mathrm{LN}}
\newcommand{\Wup}{W^{\uparrow}}
\newcommand{\Wdown}{W^{\downarrow}}

\newcommand\blfootnote[1]{%
  \begingroup
  \renewcommand\thefootnote{}\footnote{#1}%
  \addtocounter{footnote}{-1}%
  \endgroup
}

\begin{document}


{\LARGE\bfseries LARA: Lightweight Adapters in the\\[2pt]
Residual Stream for Composable Adaptation and Alignment\par}

\vspace{0.5em}
{\large\itshape Extending frozen models with composable behaviors\par}

\vspace{1.1em}

{\bfseries Pascal Ekin}\,\textsuperscript{$\dagger$}\qquad
{\bfseries Hyosun Choi}\,\textsuperscript{1}\qquad
{\bfseries Wei Jie}\,\textsuperscript{2}

\vspace{0.5em}
{\small
\textsuperscript{1}\textit{Royal Holloway, University of London, United Kingdom}\\
\textsuperscript{2}\textit{University of West London, United Kingdom}\par}

\vspace{2.2em}


\begin{spacing}{1.0}
\noindent\textbf{Abstract}\\[0.3em]
\small
We present LARA (Lightweight Additive Residual Adaptation), a method for efficient
adaptation that operates in the residual stream of a frozen model rather than in its
weights. Where LoRA adds an update of low rank to weight matrices, LARA reads the hidden
state at a small set of layers and adds a correction of low rank back to the residual
stream, leaving all base weights untouched. On a code fine-tuning task and on preference
optimization (DPO), LARA matches LoRA at equal parameter counts. Because adaptation is a
frozen base plus a residual, LARA exposes a scale $\gamma$, applied at inference, that
interpolates smoothly between base and adapted behavior, a form of graded control that
adaptation in weight space does not offer. Finally, because each behavior is a small
residual module over a shared frozen base, many behaviors can be held resident at once and
routed automatically per token. We place seven behaviors, six fine-tuned and one optimized
for preference, on one frozen 1.5B model for roughly 33~MB of overhead, against one full
model for each behavior. Because the base is untouched, behaviors are trained separately and
selected per token rather than loaded on demand, which suits hosting many behaviors, and
adding new ones, on a single model on a device.%

\blfootnote{\textsuperscript{$\dagger$}\,Correspondence: \texttt{pfekin@gmail.com}. \ Code:
\url{https://github.com/pfekin/LARA}}
\end{spacing}

\vspace{0.4em}
{\small\noindent\textbf{Keywords:} parameter-efficient fine-tuning \textbullet\ residual
stream \textbullet\ adapters \textbullet\ LoRA \textbullet\ preference optimization
\textbullet\ inference-time steering \textbullet\ on-device inference \textbullet\
per-token routing\par}

\vspace{0.8em}


\section{Introduction}

Efficient adaptation lets a large pretrained model be specialized without updating all of
its weights, at a small fraction of the cost of full fine-tuning~\cite{Houlsby2019,Hu2021}.
The dominant approach, low rank adaptation (LoRA), adds a trainable update of low rank to
selected weight matrices~\cite{Hu2021}. Methods in this family adapt a model by modifying
its weights: the learned update is folded into, or applied alongside, the matrices that
define each layer's transformation.

Weight space is not the only place adaptation can live. A transformer maintains a residual
stream, the running sum of activations that each layer reads from and writes
to~\cite{Elhage2021}. Rather than changing the weights that produce those activations, one
can leave every weight frozen and add a small learned correction to the residual stream
itself. We study this alternative in LARA (Lightweight Additive Residual Adaptation): at a
small set of layers, LARA reads the hidden state, computes a correction of low rank, and
adds it back to the stream. The base weights are never touched, so adaptation is a residual
placed beside the frozen computation rather than a change to it.

This mechanism is not new. Adding a learned module to a frozen backbone, and specifically a
module of low rank or a bottleneck on the residual path, is the family of residual adapters
and side tuning~\cite{Rebuffi2017,Houlsby2019,Zhang2020}. The closest recent instance is
H-Res~\cite{Awadhiya2026}, which adds a low rank module, initialized to zero, to the
residual stream of a frozen model and evaluates it on adaptation to a single task. We do not
claim the residual-adapter mechanism as a contribution. Our contribution is threefold.
First, we establish by comparison at equal parameters, rather than assume, that adaptation
in the residual stream matches LoRA in weight space on a code fine-tuning task and on
preference optimization (DPO); the closest prior work compares against LoRA without matching
parameter counts. Second, because the adaptation is an additive residual over an unchanged
base, its strength can be scaled at inference by a single coefficient, giving graded control
between base and adapted behavior that methods in weight space do not expose. Third, because
each behavior is a small module over a shared frozen base, many behaviors can be held
resident at once and routed automatically per token, placing a bank of behaviors, tuned and
aligned, on one frozen model for a few megabytes rather than one full model for each
behavior.

The last point is where the frozen base pays off in practice. A bank of behaviors is held
resident for a small fraction of one model's memory and selected per token, against roughly
one full copy of the model per behavior under conventional deployment. Because the base is
never modified, behaviors are trained separately and added to the bank without refitting the
model, so one frozen model can host many behaviors and acquire new ones over time. LARA
reaches this while remaining comparable in adaptation quality to weight-space adaptation and
adding a control at inference that it does not offer. Section~\ref{sec:method}
describes the method, Section~\ref{sec:exp} reports the comparison at equal parameters and
the steering and the results on many behaviors, and Section~\ref{sec:disc} discusses
limitations and relation to prior work.


\section{Method}
\label{sec:method}

\subsection{Adaptation in the residual stream}

A transformer decoder of depth $D$ maintains a residual stream: at layer $\ell$ the block
reads a hidden state $h_\ell$ and writes $h_{\ell+1}$, and the model's output is read from
$h_D$~\cite{Elhage2021}. LARA leaves every block unchanged and inserts, at a chosen subset
of layers $L$, a lightweight module that reads the hidden state and adds a correction of low
rank back to the stream. For each $\ell \in L$,
\begin{equation}
h_\ell \;\leftarrow\; h_\ell \;+\; \frac{\alpha}{r}\, \Wup\, \Wdown\, \LN(h_\ell),
\end{equation}
where $\Wdown$ is an $r \times d$ matrix, $\Wup$ is a $d \times r$ matrix, $r \ll d$ is the
rank, $\LN$ is layer normalization, and $\alpha$ is a fixed scale. The map is linear in
$\LN(h_\ell)$. There is no nonlinearity between the two projections. Following the principle
of initializing to zero used by LLaMA-Adapter~\cite{Zhang2023}, $\Wup$ is set to zero, so at
the start of training every module outputs zero and the model is exactly the base.

The base weight matrices are never modified. With the modules removed, or equivalently with
their output set to zero, the forward pass is identical to the frozen base, bit for bit.
This is the operational difference from adaptation in weight space: where LoRA replaces an
effective weight matrix $W$ with $W + (\alpha/r)BA$ and thereby alters the transformation
each layer computes, LARA leaves $W$ untouched and instead adds to the activations that flow
between layers. Adaptation is placed beside the frozen computation rather than folded into
it.

\begin{figure}[t]
\centering
\includegraphics[width=\textwidth]{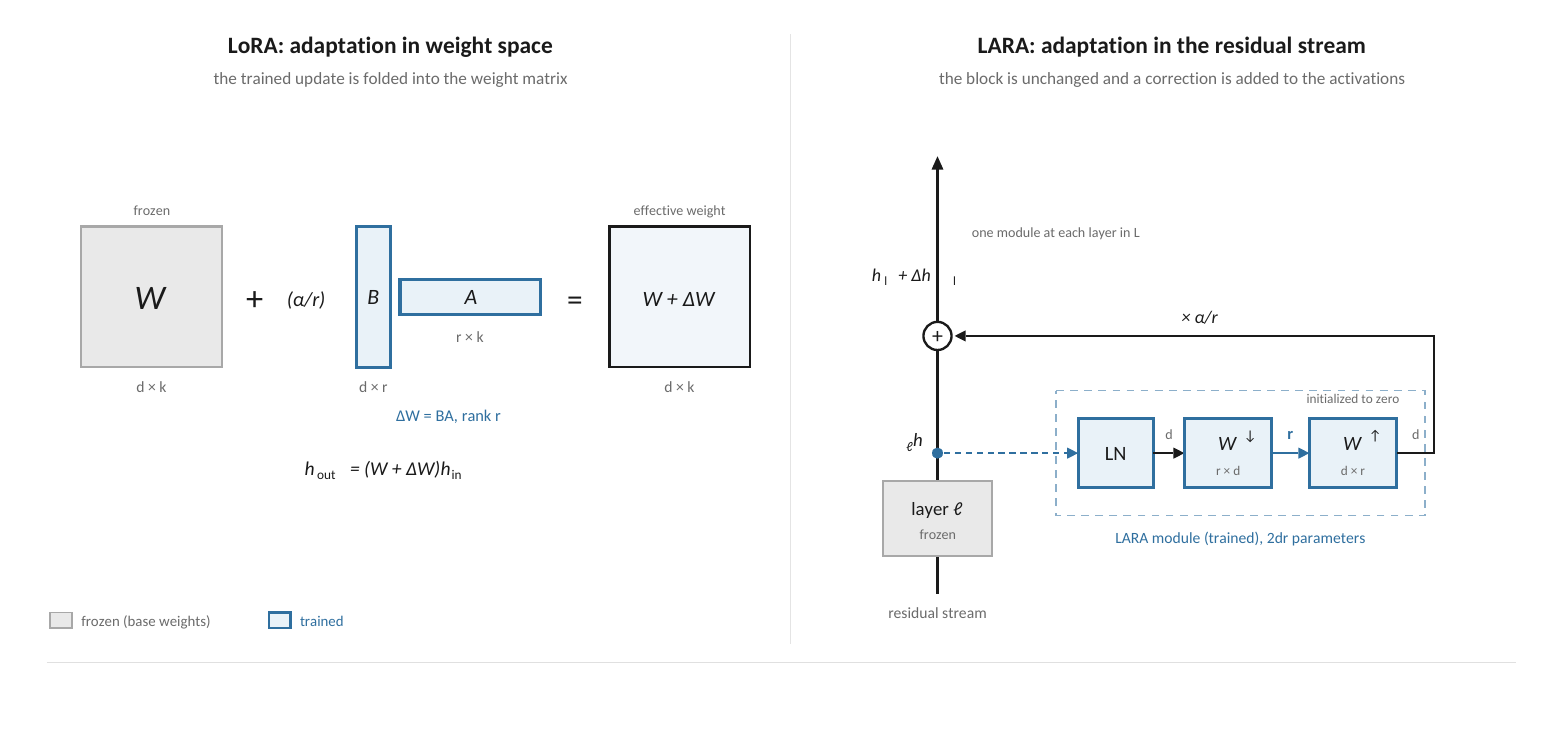}
\caption{LoRA adapts in weight space; LARA adapts in the residual stream. Left: LoRA
factorizes the update $\Delta W = BA$ of rank $r$ and folds it into the weight matrix, so
the layer computes $(W + \Delta W)h$. Right: LARA reads the hidden state $h_\ell$, projects
it down to rank $r$ and back to dimension $d$, scales by $\alpha/r$, and adds the result to
the stream, while the block itself stays frozen. Grey is frozen, blue is trained.}
\label{fig:one}
\end{figure}

\subsection{Trained parameters and accounting at equal budgets}
\label{sec:params}

Only the inserted modules are trained, and all base parameters remain frozen. Each module
holds $2dr$ projection parameters together with the affine parameters of its LayerNorm, so a
configuration over $|L|$ layers scales as approximately $2dr|L|$. In the configuration used
throughout Section~\ref{sec:exp}, with hidden size $d = 1536$, rank $r = 128$, and
$L = \{4, 8, 12, 16, 20, 24\}$, LARA trains $2{,}386{,}944$ parameters. The LoRA baseline
applies updates of rank 16 to the query and value projections across all layers, for
$2{,}179{,}072$ trainable parameters. The two budgets are within 10\%, with LARA the larger.
We report both methods at these counts and make no attempt to give either a parameter
advantage.

\subsection{Placement}

The set $L$ is the one design choice with appreciable effect on quality. Rank and scale are
comparatively insensitive within the ranges we use. Two regularities hold in our experiments
and are supported in Section~\ref{sec:exp}. Supervised fine-tuning benefits from placing
modules at several layers spanning the depth. Preference optimization is less demanding: a
single module at a middle layer reaches parity with the configuration of several layers.
Except where a single layer is stated explicitly, all reported configurations use the
placement over several layers above.

\subsection{Scaling at inference}
\label{sec:gamma}

Because each module's contribution is additive over an unchanged base, its strength can be
adjusted after training without retraining. We scale the trained correction by a coefficient
$\gamma \geq 0$ applied only at inference,
\begin{equation}
h_\ell \;\leftarrow\; h_\ell \;+\; \gamma\,\frac{\alpha}{r}\, \Wup\, \Wdown\, \LN(h_\ell),
\end{equation}
with $\gamma = 1$ recovering the trained model and $\gamma = 0$ recovering the base. During
training $\gamma = 1$ throughout, and it is introduced only as a control at evaluation.
Section~\ref{sec:exp} shows that intermediate values interpolate smoothly between base and
adapted behavior, which adaptation in weight space does not provide as a control at
inference.

\subsection{Computational cost}

A module performs two matrix products per token. The projection to rank $r$ costs $rd$
operations and the projection back to $d$ costs $dr$, counting a multiply and an add as one
operation, so a module adds $2dr$ per token and a placement over $|L|$ layers adds $2dr|L|$.
The LayerNorm adds a term linear in $d$. The map is applied at each position independently,
so the added cost is linear in the rank $r$, linear in the number of modules $|L|$, and
linear in the sequence length. It does not enter the quadratic term in attention, and the
cost of the base model is unchanged.

Because the map is linear, the operations a module adds per token equal its trainable
parameter count. At the budget of Section~\ref{sec:params} this is $2{,}386{,}944$
operations per token, against roughly $1.5 \times 10^{9}$ for the frozen base excluding the
attention scores, whose cost depends on context length. The adaptation is therefore under
$0.2\%$ of the forward pass. A bank of seven behaviors under soft routing applies every
behavior's modules and blends them, at $N \times 2dr|L|$ per token, close to $1\%$ of the
base at $N = 7$. Hard routing to one behavior applies a single set. The router is one product
of size $d \times N$ per token, $10{,}752$ operations at $d = 1536$ and $N = 7$, which is
negligible against either.

One difference from adaptation in weight space is worth stating. A LoRA update can be merged
into the weight matrix once training is finished, after which it costs nothing at inference.
LARA cannot be merged, because its correction is a function of the activation rather than a
constant offset to the weights, so it always pays the cost above. Merging is available when
a single behavior is wanted permanently. It is not available for a bank of behaviors that
must remain switchable, which is the setting of Section~\ref{sec:routing}, and against an
unmerged LoRA the two methods add the same work per token at equal parameters.


\section{Experiments}
\label{sec:exp}

We compare LARA against LoRA at the equal parameter budget of Section~\ref{sec:params}
(2.39M against 2.18M trainable, LARA the larger), on the same frozen base, Qwen2.5 1.5B
Instruct in 8 bits. Both methods are trained identically per task. Section~\ref{sec:match}
establishes that adaptation in the residual stream matches adaptation in weight space on
fine-tuning and preference optimization. Section~\ref{sec:scan} examines the scaling
coefficient at inference. Section~\ref{sec:routing} shows many behaviors held resident on one
frozen base and routed per token.

\subsection{Adaptation in the residual stream matches LoRA in weight space}
\label{sec:match}

\textit{Fine-tuning.}
We fine-tune on a code corpus\footnote{The corpus combines four public datasets,
predominantly Python with a small instruction-following remainder:
\texttt{iamtarun/python\_code\_instructions\_18k\_alpaca},
\texttt{Vezora/Tested-22k-Python-Alpaca},
\texttt{mlabonne/Evol-Instruct-Python-26k}, and
\texttt{HuggingFaceH4/ultrachat\_200k}.} (approximately 95\% Python with a small remainder of
instruction following) and evaluate perplexity on a split held out from the training
distribution and on a general instruction split (Databricks Dolly) that shares the chat
format but not the code content. Table~\ref{tab:ft} reports both methods against the frozen
base. Both reduce perplexity on the training distribution sharply, from 5.34 to 1.70 (LARA)
and 1.75 (LoRA), and both improve the instruction split over base, to 7.50 and 6.74
respectively. LARA is marginally ahead on the training distribution and LoRA marginally ahead
on the instruction split. Neither difference is large, and neither method is uniformly
better. At equal parameters, LARA reaches the same fine-tuning quality as LoRA.

\begin{table}[htbp]
\centering
\caption{Fine-tuning perplexity (lower is better) on a code corpus, evaluated on the
training distribution and on a general instruction split, against the frozen base. LARA and
LoRA at equal parameters.}
\label{tab:ft}
\begin{tabular}{lrrr}
\toprule
Method & Params & Train PPL & Instruct PPL \\
\midrule
Frozen base & --- & 5.34 & 29.54 \\
LoRA ($r{=}16$) & 2.18M & 1.75 & 6.74 \\
LARA (ours) & 2.39M & 1.70 & 7.50 \\
\bottomrule
\end{tabular}
\end{table}

\textit{Preference optimization.}
We apply DPO with length normalization on UltraFeedback preference pairs, using the frozen
base with adapters disabled as the reference, and evaluate on the held out pairs.
Table~\ref{tab:dpo} reports reward accuracy (fraction of pairs where the adapted model
prefers the chosen response), the reward margin, and the DPO loss on the held out pairs.
Both methods learn genuine preference: reward accuracy rises to 0.625 (LARA) and 0.613
(LoRA), above the frozen base's 0.55, while the base preference gap confirms the pairs are
separable prior to training. The two methods are comparable, with LARA showing a higher
reward accuracy and margin and LoRA a lower loss on those pairs. As is characteristic of
preference optimization under length normalization, both methods increase the margin between
chosen and rejected rather than raising the chosen log probability in isolation. At equal parameters, LARA matches LoRA on preference optimization as well.

\begin{table}[t]
\centering
\caption{DPO with length normalization on UltraFeedback (reward accuracy higher is better).
Both methods exceed the frozen base's reward accuracy of 0.55. LARA and LoRA at equal
parameters. The reference is the frozen base with adapters disabled, which is free.}
\label{tab:dpo}
\begin{tabular}{lrrrr}
\toprule
Method & Params & Reward acc. & Margin & DPO loss \\
\midrule
LoRA ($r{=}16$) & 2.18M & 0.613 & $+0.429$ & 0.6664 \\
LARA (ours) & 2.39M & 0.625 & $+0.600$ & 0.7627 \\
\bottomrule
\end{tabular}
\end{table}

\subsection{Scaling at inference}
\label{sec:scan}

Because each module's correction is additive over an unchanged base, the coefficient
$\gamma$ of Section~\ref{sec:gamma} rescales the adaptation at inference with no retraining.
Setting $\gamma = 0$ recovers the frozen base exactly, and $\gamma = 1$ recovers the trained
model. We evaluate $\gamma$ from 0 to 3 on the fine-tuning split from the training
distribution, for two placements: the configuration of six layers from Section~\ref{sec:match},
and a single bridge at a middle layer.

Table~\ref{tab:gamma} reports perplexity on the training distribution as $\gamma$ varies. For
both placements, perplexity moves monotonically from the base value at $\gamma = 0$ to the
trained value at $\gamma = 1$, and intermediate settings yield intermediate perplexities: the
coefficient interpolates continuously between base and adapted behavior. This provides a
single control at inference over adaptation strength that adaptation in weight space does not
expose, since a merged or applied LoRA update has a fixed magnitude at inference. The
interpolation is over perplexity, a measure of fit, and we do not claim intermediate $\gamma$
yields a separately useful behavior.

Placement governs behavior beyond $\gamma = 1$. The configuration of six layers reaches a
lower perplexity at $\gamma = 1$ (1.71 against 1.96), consistent with fine-tuning benefiting
from several layers, but degrades sharply when the trained correction is amplified: at
$\gamma = 3$ its perplexity rises to 219. The configuration of a single bridge is far more
stable under the same amplification, remaining at 5.51 at $\gamma = 3$. The two placements
thus trade adaptation strength at $\gamma = 1$ against stability under extrapolation: stacking
modules across depth improves the fit but compounds the overshoot when scaled past the
trained point, whereas a single module degrades gently. In both cases the useful range is
bounded, and $\gamma = 1$ is the trained operating point.
We do not have a full account of why amplification past $\gamma = 1$ destabilizes the six-layer configuration so abruptly while a single bridge degrades gently. The contrast suggests the effect accumulates across stacked modules rather than arising at any one layer.

\begin{table}[htbp]
\centering
\caption{Perplexity on the training distribution (lower is better) as the scale $\gamma$
varies at inference, for two LARA placements: six bridges at layers
$\{4, 8, 12, 16, 20, 24\}$ (2.39M parameters, the configuration of Section~\ref{sec:match})
and a single bridge at layer 14 (0.40M parameters). $\gamma = 0$ recovers the frozen base,
and $\gamma = 1$ is the trained model. Both interpolate from base to trained over
$\gamma \in [0,1]$. The placement of six layers fits better at $\gamma = 1$ while the
placement of a single bridge is more stable for $\gamma > 1$.}
\label{tab:gamma}
\begin{tabular}{ccc}
\toprule
$\gamma$ & Six bridges (2.39M) & Single bridge (0.40M) \\
\midrule
0.00 & 5.34 & 5.34 \\
0.50 & 2.01 & 2.34 \\
1.00 & 1.71 & 1.96 \\
1.50 & 2.49 & 2.12 \\
2.00 & 7.31 & 2.72 \\
3.00 & 219.37 & 5.51 \\
\bottomrule
\end{tabular}
\end{table}

\subsection{Many behaviors on one frozen base}
\label{sec:routing}

The additive form of LARA, which preserves the base, lets several trained behaviors remain
resident on one frozen model at once. We place seven behaviors on a single frozen Qwen2.5
1.5B Instruct: six fine-tuned (code, general instruction, math, medical, a second code set,
and summarization) and one optimized for preference with DPO. Each behavior is a separate set
of LARA modules over the same frozen base. A small linear router, about 11k parameters, reads
the base hidden state at one layer and assigns each token a distribution over the seven
behaviors, and their corrections are applied as a weighted blend, per token. Behaviors are
thus selected automatically per token at inference, rather than loaded one at a time.

Table~\ref{tab:routing} reports, for each behavior, the performance of its dedicated module
applied alone and of the same behavior under routing over the full bank, with the fraction of
dedicated performance retained. Every fine-tuned behavior retains between 0.87 and 0.99 of
its dedicated perplexity under routing, and the preference behavior retains its reward.
Behaviors from fine-tuning and from preference optimization coexist on one frozen base, with
no separate model per behavior. The router concentrates most of each token's weight on the
correct behavior for distinct domains: math and medical each route to their own behavior
above 0.94. The two code behaviors are deliberately similar, a general code corpus and a
second set of code instructions, and the router does not separate them cleanly, with the
second set assigning about a third of its weight to the first. Recovery for these behaviors
holds regardless, because the two code modules are close substitutes: a token routed to
either receives an appropriate code correction. The pair of near neighbors marks the limit of
routing precision without a matching loss in quality. 
Notably, blending does not appear to dilute confident predictions. The softness mainly helps on the ambiguous tokens that hard routing would misassign.
Under hard routing to the single top
behavior the same pattern holds, but recovery is lower, because ambiguous tokens are
committed to a single behavior rather than blended.

\begin{table}[htbp]
\centering
\caption{Seven behaviors resident on one frozen Qwen2.5 1.5B Instruct under soft routing.
For each behavior: performance of the dedicated module alone, under routing over the full
bank, the fraction of dedicated performance retained, and the mean weight the router assigns
to a behavior on its own tokens. Fine-tuned rows are perplexity, lower is better. The
preference row is reward accuracy, higher is better.}
\label{tab:routing}
\begin{tabular}{llrrrr}
\toprule
Behavior & Task & Dedicated & Routed & Recovery & Own weight \\
\midrule
code & FT & 1.77 & 1.98 & 0.89 & 0.82 \\
general & FT & 6.18 & 6.69 & 0.92 & 0.71 \\
math & FT & 1.38 & 1.41 & 0.98 & 0.96 \\
medical & FT & 7.09 & 7.14 & 0.99 & 0.94 \\
code (2nd set) & FT & 2.21 & 2.54 & 0.87 & 0.56 \\
summary & FT & 4.34 & 4.82 & 0.90 & 0.87 \\
preference & DPO & 0.640 & 0.640 & 1.00 & 0.82 \\
\bottomrule
\end{tabular}
\end{table}

Because soft routing blends behaviors per token, a behavior that cuts across domains can be
applied together with a domain behavior on the same token, which routing of one behavior per
token cannot express. We measure this for the preference behavior. On the tokens where a
domain behavior's own module leads, we record the weight assigned at the same time to the
preference behavior. On general instruction tokens the preference behavior fires alongside at
0.13, whereas on math and medical tokens it is near zero, at 0.03. The preference style is
applied selectively. It is present alongside general instruction, where the style is
relevant, and largely absent on math and medical. This measures joint activation of the
preference module on shared tokens, not a behavioral change in the generated text.
Establishing the latter would require a preference evaluation of the blended output.

The bank is compact. The frozen base occupies about 1.5~GB as 8-bit weights and 3.09~GB when
resident. Seven behaviors add 33~MB of modules and 0.02~MB of router, bringing the resident
total to 3.12~GB, against 21.6~GB for seven separate copies of the model. Holding many
behaviors on one frozen base and selecting among them per token suits settings where memory
is the binding constraint, such as deployment on a device.


\section{Discussion}
\label{sec:disc}

LARA places efficient adaptation in the residual stream of a frozen model rather than in its
weights, and reaches the quality of LoRA in weight space at equal parameters on the tasks we
study while adding two properties that follow from the additive form that preserves the base:
a scale, applied at inference, that interpolates between base and adapted behavior, and the
ability to hold many behaviors resident on one frozen model and route among them per token.

The mechanism sits within an established line of work. Adding a learned module to a frozen
backbone is the family of residual adapters and side tuning, and initializing the added
module to zero so training begins from the base is standard practice. LARA's contribution is
not the mechanism but its characterization: a comparison at equal parameters against LoRA on
both fine-tuning and preference optimization, the scaling coefficient at inference, and the
routing of many behaviors on a shared frozen base. The closest recent instance adds a module
of low rank, initialized to zero, to the residual stream of a frozen model but evaluates
adaptation to a single task without a comparison at equal parameters. The present results
supply that comparison and the composition and control that adaptation to a single task does
not exercise.

More broadly, methods for efficient adaptation can be organized by where they intervene.
LoRA and its relatives place a trainable update on the weights~\cite{Hu2021}. BitFit stays in
the same space and trains only the bias terms~\cite{BenZaken2022}. Prefix tuning and prompt
tuning leave the weights alone and prepend trainable vectors to the input or to the attention
keys and values, spending context rather than weights~\cite{Li2021,Lester2021}. Adapters
insert bottleneck modules inside the block, in sequence with the base
computation~\cite{Houlsby2019}. (IA)\textsuperscript{3} acts on activations, rescaling keys,
values and feedforward activations by learned vectors, which is multiplicative where LARA is
additive~\cite{Liu2022}. The methods closest to LARA leave the block untouched and act on the
residual path. Side tuning trains an additive side network~\cite{Zhang2020}, ladder side
tuning routes intermediate activations into a separate side network through shortcut
connections~\cite{Sung2022}, and H-Res adds a module of low rank, initialized to zero, to the
stream of a frozen language model~\cite{Awadhiya2026}. LARA belongs to this last group.

Composing and serving several adapted behaviors has its own line of work. AdapterFusion
learns to combine separately trained adapters for a target task~\cite{Pfeiffer2021}, and
S-LoRA serves many LoRA adapters from one base by paging adapter weights between main memory
and the accelerator as requests arrive~\cite{Sheng2023}. Both share the premise that one
frozen base can support many behaviors. They differ in what the composition is for.
AdapterFusion composes adapters into a single task, and S-LoRA switches between adapters per
request in a datacenter. LARA selects among resident behaviors per token, on one device, with
the whole bank in memory at once.

Routing tokens to modules also has a relative in mixture-of-experts models, which route each
token to a small number of expert layers. The aim differs. A mixture of experts routes to add
capacity to one model, and it selects few experts per token because each is a full
feedforward block. The behaviors here are residual modules of a few megabytes, light enough
that the router can apply all of them and blend by weight rather than commit to a hard
selection, and the bank makes several behaviors available on a base that is not made larger.

Because the base is never modified, behaviors are trained independently of one another and
composed after the fact. A behavior can be added to the bank, or an existing one replaced,
without retraining the base or the other behaviors. This suits a setting where behaviors
accumulate and are maintained separately over time, closer to a set of modules over a fixed
base than to a single model refitted whenever requirements change. We do not evaluate this
workflow and note it as a property of the design rather than a measured result.

Several limitations bound the claims made here. The fine-tuning comparison uses a single code
corpus, so parity is established for a code fine-tuning task rather than fine-tuning in
general. Other domains and larger instruction mixtures remain to be tested. The scaling at
inference and the measurement of joint application both concern perplexity and the level of
weights: $\gamma$ interpolates a measure of fit, and the joint application measures where the
preference module is active on shared tokens, not whether the blended text is better aligned
to preference. Confirming a behavioral effect would require downstream evaluation of the
scaled and blended outputs. All results are on one 1.5B base model in 8 bits. Scaling
behavior across model sizes is not examined here. The empirical comparison is against LoRA
alone. The placement of LARA among the other families above is positioning rather than
measurement, and baselines at equal parameters against them are left to future work. Finally,
routing precision is limited for behaviors that are close in content, as the two code
behaviors show, though quality is retained when the confusable behaviors are near
substitutes.

These results suggest that adaptation, along with its scaling and composition, can be handled
together in the residual stream of a single frozen model. Routing many behaviors per token at
a cost of megabytes is the property most specific to this setting, and points toward
deployment on a device or under limited memory where many behaviors must share one model.
Whether adaptation in the residual stream retains parity with methods in weight space at
larger scale, and whether the scale at inference and the blending per token yield
controllable behavioral differences and not only perplexity differences, are the natural next
questions.



\appendix
\section{Implementation details}

\textbf{Base model.} Qwen2.5 1.5B Instruct, loaded in 8 bits, hidden size $d = 1536$, 28
layers. All experiments were run on single-GPU Google Colab instances, the notebooks to
reproduce them are in the repository.

\textbf{LARA.} Rank $r = 128$, scale $\alpha = 128$, linear projection with a trained
LayerNorm on the module input, the up projection initialized to zero. Fine-tuning placement
$L = \{4, 8, 12, 16, 20, 24\}$ ($2{,}386{,}944$ trainable parameters). The single-layer
configuration of Section~\ref{sec:scan} places one module at layer 14 ($397{,}824$
parameters). The scale $\gamma$, applied at inference, is 1 during training.

\textbf{LoRA baseline.} Rank 16 on the query and value projections across all layers,
$\alpha = 2r$, dropout 0.05 ($2{,}179{,}072$ trainable parameters).

\textbf{Fine-tuning.} Code corpus of roughly 95\% Python with a small remainder of
instruction following, formatted for chat. Training runs 600 steps, learning rate $2\times
10^{-4}$, gradient accumulation 4, maximum sequence length 512. Perplexity is measured on a
split held out from the training distribution. The general instruction split is Databricks
Dolly.

\textbf{Preference optimization.} DPO with length normalization on UltraFeedback (binarized),
$\beta = 5.0$, NLL regularization weight 0.1, 60 steps, learning rate $2\times 10^{-4}$,
gradient accumulation 16, 512 training pairs and 256 evaluation pairs, maximum prompt and
response length 128. The reference is the frozen base with adapters disabled.

\textbf{Routing.} Seven behaviors share one frozen base. A linear router (about 11k
parameters) reads the base hidden state at one layer and produces a per-token distribution
over behaviors, applied as a soft weighted blend of the behaviors' corrections. Footprint:
33~MB of behavior modules and 0.02~MB of router over a 3.09~GB base.

\textbf{Code.} Configuration and scripts to reproduce these experiments are available at
\url{https://github.com/pfekin/LARA}.


\begin{thebibliography}{99}
\setlength{\itemsep}{2pt}

\bibitem{Awadhiya2026}
Awadhiya, K. Parallel Manifold Steering (H-Res).
\href{https://arxiv.org/abs/2606.24396}{arXiv:2606.24396}, 2026.

\bibitem{BenZaken2022}
Ben Zaken, E., Goldberg, Y., and Ravfogel, S. BitFit: Simple Parameter-efficient Fine-tuning
for Transformer-based Masked Language-models. In \textit{Proceedings of the 60th Annual
Meeting of the Association for Computational Linguistics (Volume 2: Short Papers)}, 2022.
\href{https://arxiv.org/abs/2106.10199}{arXiv:2106.10199}.

\bibitem{Elhage2021}
Elhage, N., Nanda, N., Olsson, C., et al. A Mathematical Framework for Transformer Circuits.
\textit{Transformer Circuits Thread}, 2021.
\url{https://transformer-circuits.pub/2021/framework/index.html}

\bibitem{Houlsby2019}
Houlsby, N., Giurgiu, A., Jastrzebski, S., Morrone, B., de Laroussilhe, Q., Gesmundo, A.,
Attariyan, M., and Gelly, S. Parameter-Efficient Transfer Learning for NLP. In
\textit{International Conference on Machine Learning (ICML)}, 2019.
\href{https://arxiv.org/abs/1902.00751}{arXiv:1902.00751}.

\bibitem{Hu2021}
Hu, E. J., Shen, Y., Wallis, P., Allen-Zhu, Z., Li, Y., Wang, S., Wang, L., and Chen, W.
LoRA: Low-Rank Adaptation of Large Language Models.
\href{https://arxiv.org/abs/2106.09685}{arXiv:2106.09685}, 2021.

\bibitem{Lester2021}
Lester, B., Al-Rfou, R., and Constant, N. The Power of Scale for Parameter-Efficient Prompt
Tuning. \href{https://arxiv.org/abs/2104.08691}{arXiv:2104.08691}, 2021.

\bibitem{Li2021}
Li, X. L. and Liang, P. Prefix-Tuning: Optimizing Continuous Prompts for Generation. In
\textit{Proceedings of the 59th Annual Meeting of the Association for Computational
Linguistics (ACL)}, 2021. \href{https://arxiv.org/abs/2101.00190}{arXiv:2101.00190}.

\bibitem{Liu2022}
Liu, H., Tam, D., Muqeeth, M., Mohta, J., Huang, T., Bansal, M., and Raffel, C. Few-Shot
Parameter-Efficient Fine-Tuning is Better and Cheaper than In-Context Learning.
\href{https://arxiv.org/abs/2205.05638}{arXiv:2205.05638}, 2022.

\bibitem{Pfeiffer2021}
Pfeiffer, J., Kamath, A., R\"uckl\'e, A., Cho, K., and Gurevych, I. AdapterFusion:
Non-Destructive Task Composition for Transfer Learning. In \textit{Proceedings of the 16th
Conference of the European Chapter of the Association for Computational Linguistics (EACL)},
2021. \href{https://arxiv.org/abs/2005.00247}{arXiv:2005.00247}.

\bibitem{Rebuffi2017}
Rebuffi, S.-A., Bilen, H., and Vedaldi, A. Learning Multiple Visual Domains with Residual
Adapters. In \textit{Advances in Neural Information Processing Systems (NeurIPS)}, 2017.
\href{https://arxiv.org/abs/1705.08045}{arXiv:1705.08045}.

\bibitem{Sheng2023}
Sheng, Y., Cao, S., Li, D., Hooper, C., Lee, N., Yang, S., Chou, C., Zhu, B., Zheng, L.,
Keutzer, K., Gonzalez, J. E., and Stoica, I. S-LoRA: Serving Thousands of Concurrent LoRA
Adapters. \href{https://arxiv.org/abs/2311.03285}{arXiv:2311.03285}, 2023.

\bibitem{Sung2022}
Sung, Y.-L., Cho, J., and Bansal, M. LST: Ladder Side-Tuning for Parameter and Memory
Efficient Transfer Learning. In \textit{Advances in Neural Information Processing Systems
(NeurIPS)}, 2022. \href{https://arxiv.org/abs/2206.06522}{arXiv:2206.06522}.

\bibitem{Zhang2020}
Zhang, J. O., Sax, A., Zamir, A., Guibas, L., and Malik, J. Side-Tuning: A Baseline for
Network Adaptation via Additive Side Networks. In \textit{European Conference on Computer
Vision (ECCV)}, 2020. \href{https://arxiv.org/abs/1912.13503}{arXiv:1912.13503}.

\bibitem{Zhang2023}
Zhang, R., Han, J., Liu, C., Gao, P., Zhou, A., Hu, X., Yan, S., Lu, P., Li, H., and Qiao, Y.
LLaMA-Adapter: Efficient Fine-tuning of Language Models with Zero-init Attention.
\href{https://arxiv.org/abs/2303.16199}{arXiv:2303.16199}, 2023.

\end{thebibliography}
\end{document}